# RGB-D-based Framework to Acquire, Visualize and Measure the Human Body for Dietetic Treatments †

**Andres Fuster-Guillo [1],\*, Jorge Azorin-Lopez [1], Marcelo Saval-Calvo [1],
Juan Miguel Castillo-Zaragoza [1], Nahuel Garcia-D'Urso [1] and Robert B. Fisher [2]**

[1] Department of Computer Technology. University of Alicante. Carretera Sant Vicent del Raspeig s/n, 03690, Spain; jazorin@ua.es (J.A.L.); msaval@dtic.ua.es (M.S.C.); jmcastillo@ua.es (J.M.C.Z.); negd1@alu.ua.es (N.G.D.)

[2] School of Informatics. University of Edinburgh. 10 Crichton St, Edinburgh EH8 9AB, UK; rbf@inf.ed.ac.uk

\* Correspondence: fuster@ua.es; Tel.: +34-965-903-400

†This paper is an extended version of our paper published in "3D Technologies to Acquire and Visualize the Human Body for Improving Dietetic Treatment" Proceedings of 13th International Conference on Ubiquitous Computing and Ambient Intelligence UCAmI, 2019, doi: 10.3390/proceedings2019031053.

**Abstract:** This research aims to improve dietetic-nutritional treatment using state-of-the-art RGB-D sensors and virtual reality (VR) technology. Recent studies show that adherence to treatment can be improved using multimedia technologies. However, there are few studies using 3D data and VR technologies for this purpose. On the other hand, obtaining 3D measurements of the human body and analyzing them over time (4D) in patients undergoing dietary treatment is a challenging field. The main contribution of the work is to provide a framework to study the effect of 4D body model visualization on adherence to obesity treatment. The system can obtain a complete 3D model of a body using low-cost technology, allowing future straightforward transference with sufficient accuracy and realistic visualization, enabling the analysis of the evolution (4D) of the shape during the treatment of obesity. The 3D body models will be used for studying the effect of visualization on adherence to obesity treatment using 2D and VR devices. Moreover, we will use the acquired 3D models to obtain measurements of the body. An analysis of the accuracy of the proposed methods for obtaining measurements with both synthetic and real objects has been carried out.

**Keywords:** obesity; adherence; 3D vision; RGB-D sensors; virtual reality; human body measures; dietetic treatment

## 1. Introduction

The prevalence of overweight and obesity has increased globally, tripling over the last three decades in the countries of the European Union. Overweight and obesity contribute to the emergence of chronic diseases (hypertension, type II diabetes, cancer, etc.) and the development of neurodegenerative pathologies (Alzheimer's or dementias) [1,2]. The treatment and follow-up care required by these patients has a high impact on the costs of health services [3–5]. Adherence to the treatment of obesity has been considered one of the factors causing the failure of intervention programs [6–8]. Given this evidence, improving adherence would contribute to the outcome of treatments and their maintenance over time, leading to lower health costs.

Some authors have suggested that nutritional interventions that reinforce the follow-up of therapies through the use of technologies achieve beneficial effects over time [8–10]. The results show how adherence to treatment can be increased by incorporating techniques based on the use of (2D) images of the patient's evolution in dietary treatment that enhance their cognitive experience [11]. However, to the best of our knowledge, there are no studies that take advantage of and quantify the potential of using realistic 3D images and virtual reality techniques to reinforce adherence to treatment. This work provides the necessary framework for specialists to address these types of studies [12].

Measuring the volume of the human body with the aim of analyzing fat concentration as a symptom of overweight and obesity is a task often addressed in the health sector with traditional techniques and single-dimensional measurements, such as waist size, weight, etc. The study of anthropometric measurements and their variation over time in relation to fat accumulation presents multidisciplinary challenges of interest in the fields of information technology and health. The use of RGB-D devices can help to address the tasks of 3D scanning human bodies and later automatically obtaining 3D and 4D measurements of the selected human body volumes, with the inherent advantages of this kind of consumer-oriented technology [13].

Nowadays, 3D modelling of the human body is transforming our ability to accurately measure and visualize bodies, showing great potential for health applications from epidemiology to diagnosis or patient monitoring [14,15]. Recent works [16,17] deal with the acquisition of images of human bodies from RGB-D cameras and video sequences, providing 3D models with texture and avatars, but they do not achieve the accuracy necessary for health applications. There are systems on the market for body model acquisition: Fit3D [18] uses a single camera but needs a complex and expensive device; Body Labs, acquired by Amazon, has also developed technology to capture the human body in three dimensions, but it does not consider any time parameter; Naked Labs [19] provides body visualization and tracking functions for the fitness market.

There are different types of 3D sensors with different characteristics that can be used to capture the human body. Devices based mainly on lasers, such as lidar, have high accuracy but they only provide depth information and do not provide color data. Stereo sensors use two color cameras to infer the depth. High-accuracy stereo systems are usually expensive, bulky, and of difficult portability since both cameras must be re-calibrated every time the system is moved. RGB-D cameras (such as the Microsoft Kinect or Intel RealSense) integrate color and depth in a single device, and they use different technologies to estimate the depth (structured light, ToF, active stereo). The characteristics of these RGB-D devices, including accuracy, portability, capture frequency, etc., are causing their popularization and integration in mobile consumer devices [20]. For these reasons, the research presented here makes use of these devices to capture 3D models. Moreover, the use of RGB-D device networks is proposed to meet the required quality levels for 3D representations.

Virtual reality has also experienced substantial growth in recent years. This technology simulates realistic 3D interactive environments using HMDs (head-mounted displays), also known as virtual reality glasses. At first, these devices had high prices due to their complexity. However, recent developments such as "Google Cardboard" allowed any smartphone to be converted into a HMD with very low costs. Currently, there is much research using the potential of virtual reality. The immersive experience provided by virtual reality stands out, improving concentration in the training process [21]. It is hoped that the use of these technologies can be effective in improving adherence to the treatment of overweight patients [10].

Classical treatments for obese patients have shown limited effectiveness in resolving chronicity due to the lack of adherence and can be improved using technology. The use of technologies for 3D reconstruction of the human body is sufficiently mature. Virtual reality is also evolving remarkably, finding low-cost systems and successful application experiences in different fields. In this context, and given the fact that there have not been previous works that have combined RGB-D acquisition devices and virtual reality to analyze body shape changes over time (4D) to improve adherence to obesity treatments, we propose an RGB-D-based framework to acquire, visualize and measure the human body for dietetic treatments.

There are important scientific challenges to the development of computational methods for the study of changes in the shape of the human body using 3D/4D vision techniques to improve obesity treatment processes. In this paper, we present a system capable of obtaining a 3D model of the body with color texture representation and a realistic visualization for 2D and virtual reality devices (see Figure 1) in order to be used for measuring the body and studying the effect of 3D technologies on adherence to obesity treatment.

For the development of this research, the following specific objectives have been addressed:

- **Obtaining the 3D/4D model**: Complete (i.e., from all sides) 3D acquisition of the human body using a low cost RGB-D camera network, obtaining the 3D geometric model and the texture representation of sequences of bodies over time (4D).
- **Visualization of the 3D body**: From the 3D models captured over time, realistic visualizations of the body evolution are generated using virtual reality.
- **Measuring selected volumes of the human body:** Selection of different parts of the human body to obtain 1D, 2D and 3D measurements.

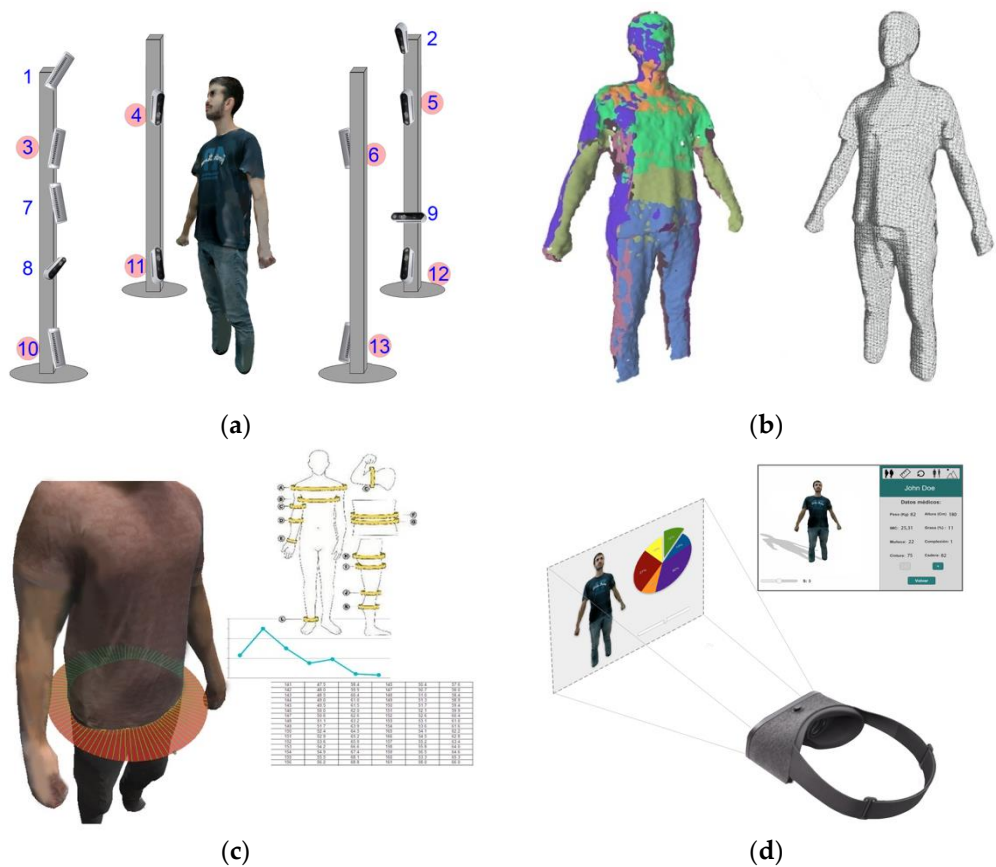

**Figure 1.** Proposed framework system based on an RGB-D camera network (**a**). Body models: point clouds and mesh (**b**). 3D body measurement (**c**) and textured representation with virtual reality (VR) devices (**d**).

The rest of the paper is organized as follows: Section 2 describes the methods and materials proposed to undertake the objectives of the research: the acquisition system for reconstructing the human body from RGB-D cameras, the system for visualization of the human body and the body measuring methods. Section 3 details the results of the proposed methods: the visualization and body measurement results. Finally, Section 4 presents the conclusions and summarizes the nutritional intervention plans and their impact and expected results to be obtained using the system.

## 2. Materials and Methods

In this section, we present the different methods and materials used to achieve the objectives previously mentioned. First, we explain the methods to acquire the realistic reconstruction of the human body using multiple RGB-D devices. The second subsection describes the tools developed for the 4D visualization of the human body using VR devices. Finally, the third subsection describes the methods developed to estimate the 3D body measurements.

*2.1. 3D Reconstruction of the Human Body from Multiple RGB-D Views*

In order to obtain the 3D model of the human body, a network of RGB-D cameras was placed in the best positions to maximize capture quality. The distance from cameras to subject was experimentally estimated to be the best in the range of 0.8—1 m, and the distribution was set to cover the whole body. The setup is composed by 13 Intel RealSense RGB-D cameras located in a cabin with 4 aluminum masts and privacy panels of 2200 × 800mm distributed around the capture area. Figure 2 shows the setup to capture a human body (a) and an example of depth and RGB color images obtained by the system (b). The camera system is organized in 5 levels of height Figure 1a: super-high (220 cm) cameras (1,2); high (180 cm) cameras (3,4,5,6); an intermediate (120 cm) camera (7); low (72 cm) cameras (8,9); and super-low (41 cm) cameras (10,11,12,13). An 8-camera version (highlighted numbers in Figure 1) has also been tested to compare measurement accuracy results, with the cameras on the levels high (180 cm) cameras (3,4,5,6) and super-low (41 cm) cameras (10,11,12,13).

The pipeline used to obtain the 3D textured model from different RGB-D sensors has five stages, as shown in Figure 4: acquisition, pre-processing, view fusion, mesh generation and texture projection. The camera network needs to be pre-calibrated, both intrinsically for each camera and multi-camera, to estimate the relative position of each sensor in the set. This calibration phase is previous to the acquisition (not in the pipeline) but crucial for the quality of the results.

The following sections describe the different methods that are part of the previously mentioned pipeline in Figure 4. Some of these are well-known methods in the state of the art and are simply referenced. While the main contribution of the work is to provide a framework to study the effect of 4D body model visualization on adherence to obesity treatment, at the methodological level, the main proposals are the calibration methods based on 3D markers and the methods for obtaining body measurements, which will be explained in more detail. Section 2.1.1 explains the calibration method and Section 2.3 describes the body measuring methods.

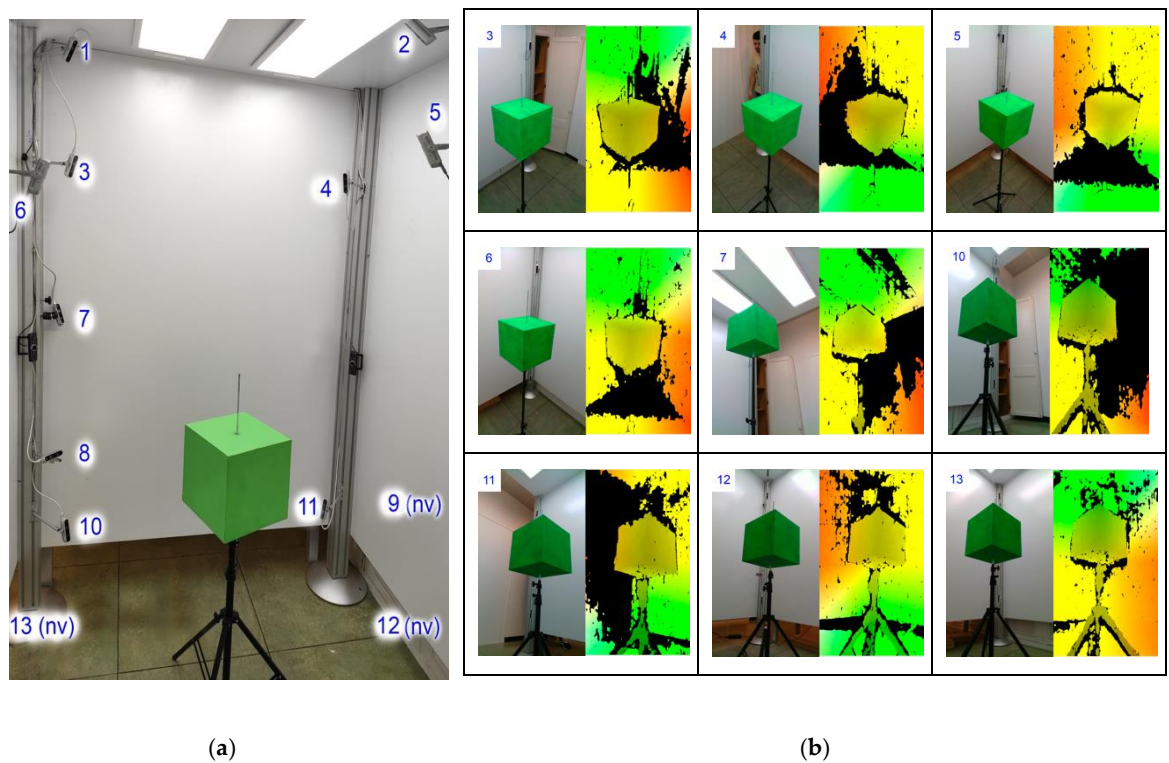

(**a**)          (**b**)

**Figure 2.** (**a**) Experimental set-up 13 RGB-D cameras. "nv" means a camera not visible in this image. (**b**) Acquisition results for the cameras (color and depth images). The green cube has been used for extrinsic calibration. Images from high, intermediate and super-low levels are shown.

2.1.1. Calibration

To correct the distortions of the images caused by the lenses, an intrinsic calibration was carried out using the provided Intel RealSense SDK. The calibration method requires a minimum of six samples of the given chessboard marker from different points distributed in the capture area.

Since we were using a network of RGB-D cameras, it was necessary to carry out an extrinsic multi-camera calibration to unify the data from each camera into the same coordinate space. To obtain the transformation matrices for each camera, we carried out an extrinsic calibration based on 3D markers, specifically using a cube as 3D marker. As presented in Figure 2b, the marker is placed in different positions of the capturing space and acquired in each position from all the cameras. With this information, an algorithm uses the known shape of the cube to robustly find correspondences, estimates the transformations to align the marker and, hence, aligns the cameras [22,23].

First of all, the cube is reconstructed by a method able to simultaneously fit a set of planar patches considering geometric constraints among them. In this case, the constraints are that all visible planar patches by the camera must be orthogonal. It achieves the model by three steps: a clustering, a regression and a reassignment (see Figure 3). A clustering is initially done to obtain 3 clusters referring to the cube planes (three is the maximum number of planes of a cube that could be seen by a camera) from the set of points, and then normal vectors are calculated from the neighborhood of each point. Secondly, a regression step, using the points of the clusters assigned, fits 3D planes of a cube model such that the orthogonality constraints are satisfied. The last step, called the reassignment step, challenges the membership of the points to the clusters. Regression and reassignment steps are repeated until no (or only a few) reassignments are performed. With this process, we obtain those captures that are valid.

Once the models of the cube have been estimated in different positions for all cameras, their centers are calculated. The centers allow us to obtain a set of representative points with which to calculate the correspondences between cameras and the corresponding transformations. The calculation of the transformations between cameras is carried out with a Procrustes analysis. In order to avoid outliers and to make it robust, RANSAC [24] is applied. This process allows the selection of a set of views that minimize the error in the calculated transformation. This is the one that minimizes the angles between the normal vectors of the patches and the distance between the centers. This process is first carried out for each row of cameras independently, as they can see common planes, and, after the rows are aligned, each using the other cameras per row that are in the same mast.

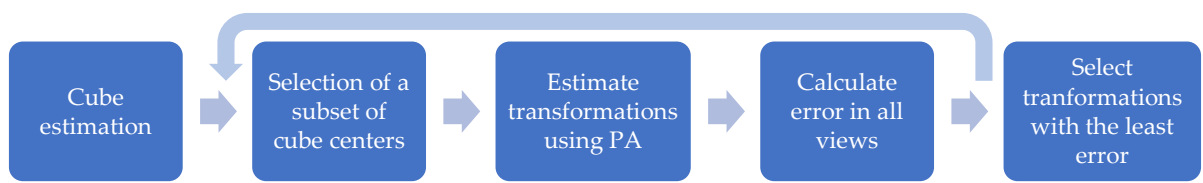

**Figure 3.** Pipeline of the calibration method based on 3D markers.

2.1.2. 3D Model Generation

As stated above, the sensor subsystem was composed of thirteen Intel RealSense RGB-D D435 cameras. Intel's SDK for RealSense was used for the development of the acquisition software. Once the system has been calibrated, the acquisition process requires the synchronization of all cameras, so that the captured data all come from the same time. Semaphore management was used to address the synchronization. The semaphores act in a similar way to a barrier and assure that all threads begin the acquisition practically at the same time, and they do not finish the acquisition of the frame until the rest of the cameras have finished. The synchronization of the cameras allows the acquisition of images with very short time delays so that the movement errors between frames are less than the errors of the camera itself.

Once the acquisition has been performed, the framework includes a pre-processing stage where the noisy point clouds are improved (see Figure 4b). First, the point cloud was truncated in the z-axis (depth) to remove the points that were beyond the center of the capture area as those tend be more affected by the noise. After that, three filters were applied: median, bilateral and statistical outlier removal (SOR). Median filters are able to reduce noise and are very efficient in terms of processing time, as they require a single pass over the cloud [25]. Bilateral filters smooth edges and areas with high curvatures while preserving sharpness using a non-linear combination of values from nearby areas [26]. SOR filters remove edge noise and outliers using neighborhood statistics [27]. Finally, the normal vector for each point in the cloud was calculated by estimating the normal of a plane tangent to the surface [28].

The third phase of the process is the data fusion. In order to align the different point clouds in a single 3D coordinate system, the transformation matrices T obtained from the extrinsic calibration were applied to the data extracted from each camera (see Figure 4c). Each camera has its own reference system with the origin (0,0,0) located in its own center. With the extrinsic parameters from the calibration, we assumed one camera as a reference and transformed the rest of the point clouds to this one to obtain a unified dataset [29].

With the complete 3D point cloud, a dense mesh representation is calculated. Different methods such as greedy projection or marching cubes were tested, but the best result was obtained using the Poisson surface algorithm [30]. It is possible to reconstruct a triangle mesh from a set of oriented 3D points by solving a Poisson system—that is, by solving a 3D Laplacian system with positional value constraints. This method approaches the problem of surface reconstruction using an implicit function framework. It computes a 3D indicator function that returns as greater than 0 the points inside the surface and as less than 0 the points outside it. This function can be found because there is a relationship between the orientation of the points and the function itself. Specifically, the gradient of the function to be found is a vector field with a value of 0 in all points except those close to the surface, which takes the value of the surface normal oriented towards the interior (see Figure 4d). After that, the reconstructed surface is obtained by extracting an appropriate isosurface.

In order to create a realistic visualization, the mesh is textured using projection techniques. The method proposed by Callieri et al. [31] was used to carry out the raster projection and texture generation. This method generates parameters of the mesh in relation to its vertices and generates the texture based on the projection of the different images considering the position and orientation of the different cameras. Figure 4e shows the same body represented by the mesh with the projected texture. A demo video of the whole process of the 3D model generation can be found in [32].

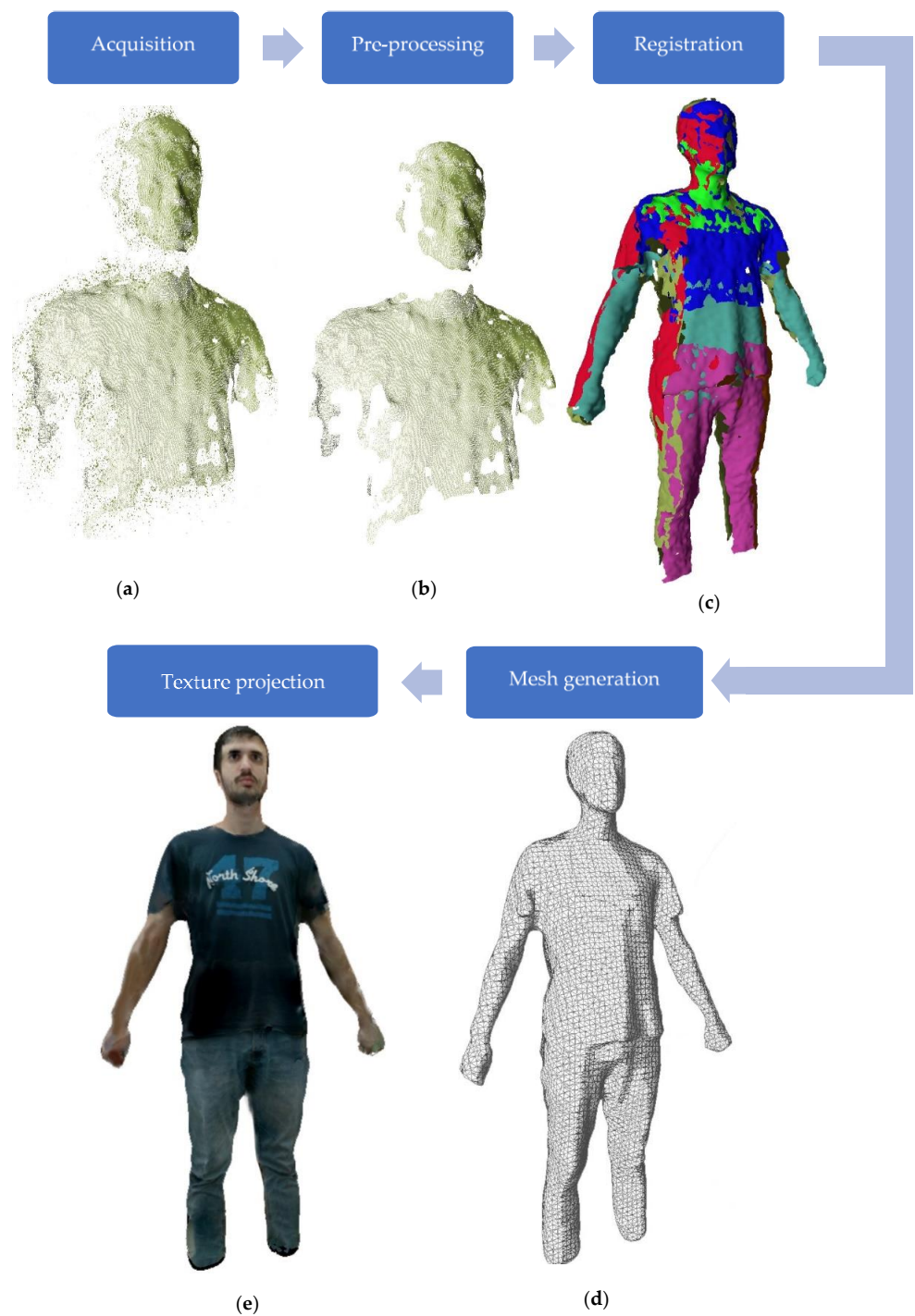

**Figure 4.** Pipeline of 3D body reconstruction. The system is able to acquire several images from cameras (**a**) that are preprocessed in order to improve the quality of the acquisition (**b**). The set of points are registered in a single coordinate system (**c**). Finally, in order to obtain the dense 3D model of the body, the 3D points are converted into a mesh (**e**) and the color images that were captured synchronously with the point cloud are texture-mapped onto the mesh (**d**).

*2.2. Visualization of the Human Body Using Virtual Reality for Obesity Treatment Improvement*

The second objective of this work is to provide a visualization system for presenting the generated 3D models. This system allows interaction with the acquisition subsystem, management of patient data, and a realistic visualization of the human body models over time. This system would be used both by the medical specialist to assist research in the field of obesity treatment and by the patient to improve their adherence to treatment by self-visualizing their improvement. Thus, the system is composed of two subsystems: the specialist 4D visualization system for obesity treatment and the virtual reality visualization system. Both parts, explained in detail next, were developed using Unity [33].

2.2.1. Specialist 4D Image Visualization System for Obesity Treatment

The visualization system allows the medical specialist several options. From the system, they can perform a scan of the patient's body since the visualization system communicates with the acquisition system. In addition, the system provides the possibility to connect with the virtual reality subsystem so the patient can see themselves with the VR glasses while the specialist is interacting with the virtual model, visualizing different options.

The system provides different options to visualize the body models generated. For example, the possibility of changing the patient's point of view. In Figure 5, we can see two different views of the same session 3 of the patient and his medical data.

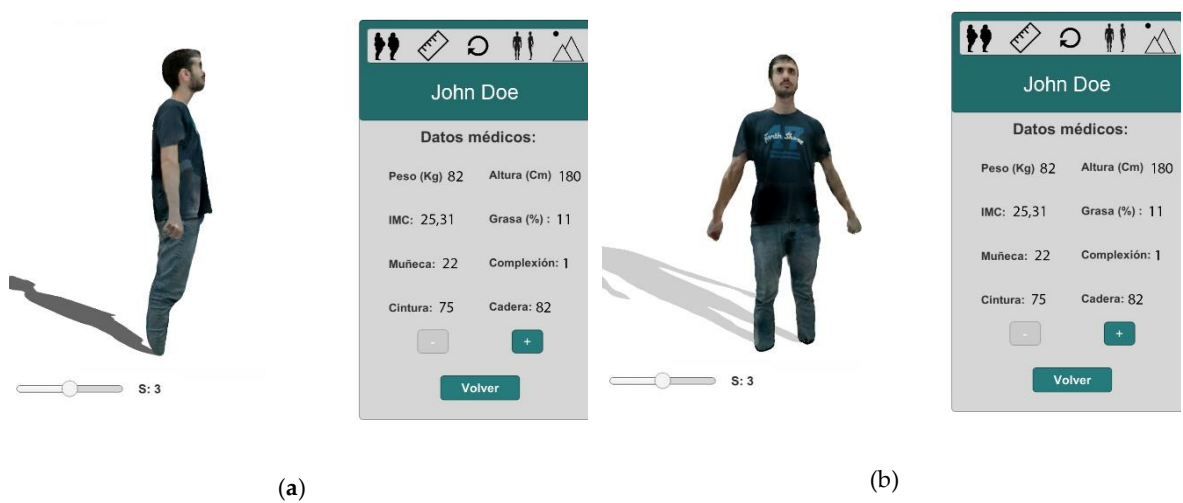

(a)      (b)

**Figure 5.** (**a**) Model visualization in the third session and its medical data. Lateral view. (**b**) Same model in the same session. Frontal view.

The system allows the visualization of the sequence of sessions for the patient to analyze its evolution over time, providing a 4D visualization. In Figure 6, we can see the scroll bar that allows navigation between sessions. In this case, it is located in session 1.

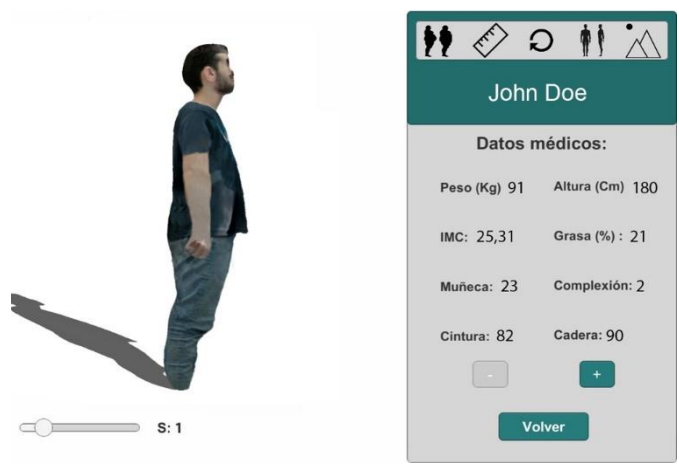

**Figure 6.** 4D visualization of a sequence of 3D models, fat session side view.

In Figure 7, we can see two superposed images of the patient to compare the two selected sessions. The two scroll bars allow selection of the sessions to be compared. One of the superposed images has the texture projected while the other is a translucent image.

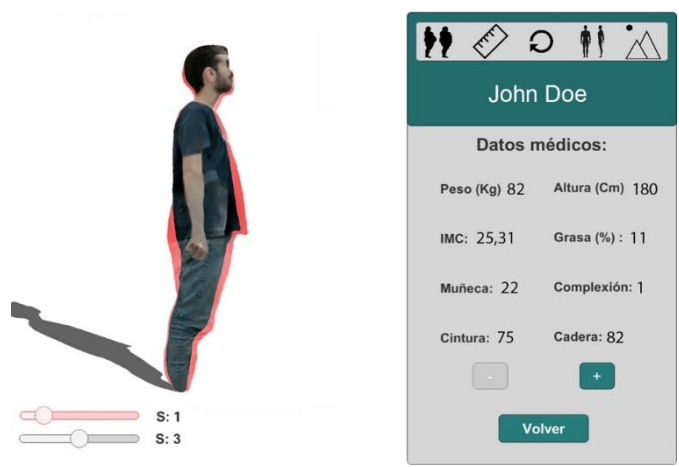

**Figure 7.** 4D visualization of a sequence of 3D models, comparing sessions.

2.2.2. Virtual Reality System

The second subsystem is composed of a virtual reality (VR) system contained in a mobile application that allows patients to see their progress in a more immersive way with the goal of improving adherence to treatments.

Two main functionalities were developed for this subsystem. First, the VR system was synchronized in real time with the computer specialist visualization system, allowing interaction between both systems. Second, the movement of the user's head was transmitted to the rotation of the 3D model being visualized (Figure 8).

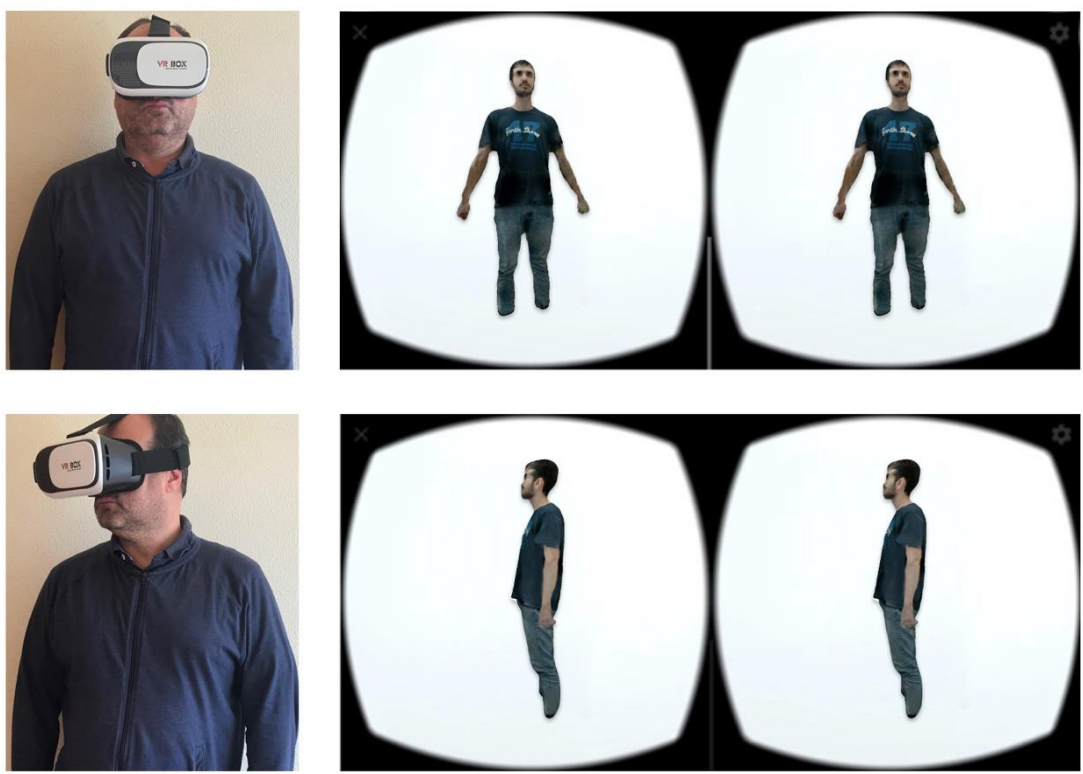

**Figure 8.** Transmission of motion to the 3D model with head positions (left) and rotated stereo images (right).

*2.3. Body Measuring Methods*

From the mesh-based 3D models generated by the pipeline described in Section 2.1, methods are proposed below for obtaining perimeter, area and volume measurements of the selected 3D model sections. Given that the software developed for the visualization makes use of the Unity graphic engine, its ray simulation system has been used as the core tool for the calculation of the points in space that determine the sections of the 3D model to be measured.

The following sections detail the method for obtaining perimetral measurements based on Unity, the method for the positioning of circles that allows the selection of the sections to be measured, and, finally, the method for obtaining measurements of areas and volumes.

Since the system has been designed to be used by dietetic specialists, the selection of the perimeters and volumes to be measured must be done by the specialist and not automatically. An interface has been designed to allow a circle to be placed anywhere in the space, so the specialist should indicate the height of the 3D model of the body to be measured. In addition, the circle can be angled to be properly oriented for measuring body parts such as arms. Finally, the size of the circle is automatically adjusted to the surrounding figure. In Figure 9a, a circle can be seen sectioning the waist part of the body. For the selection of volumes, a similar scheme has been chosen in which a cylinder is located and sized around the part of the body whose volume is to be measured.

2.3.1. Perimetral Measurement Method

A method based on ray simulation has been used to measure the perimeter. This method consists of placing a circle cutting the entire perimeter of the part to be measured. Thousands of coplanar orbiting rays are simulated on the chord all around the circle. The coplanar rays are launched along the contour to be measured, perpendicularly to tangent lines, storing the points of collision with the mesh. This process makes it possible to obtain the set of points of the mesh on the perimeter at the height of the circle (Figure 9a). As the rays are launched, the points impacted are stored and the distance between two consecutive points are accumulated. The sum of all the distances forms the desired perimeter. The number of points impacted is related to the number of rays projected. The greater the number of rays, the higher the accuracy and the computational cost, as we can see in the experimental section. This allows the number of points to be adjusted according to the accuracy required.

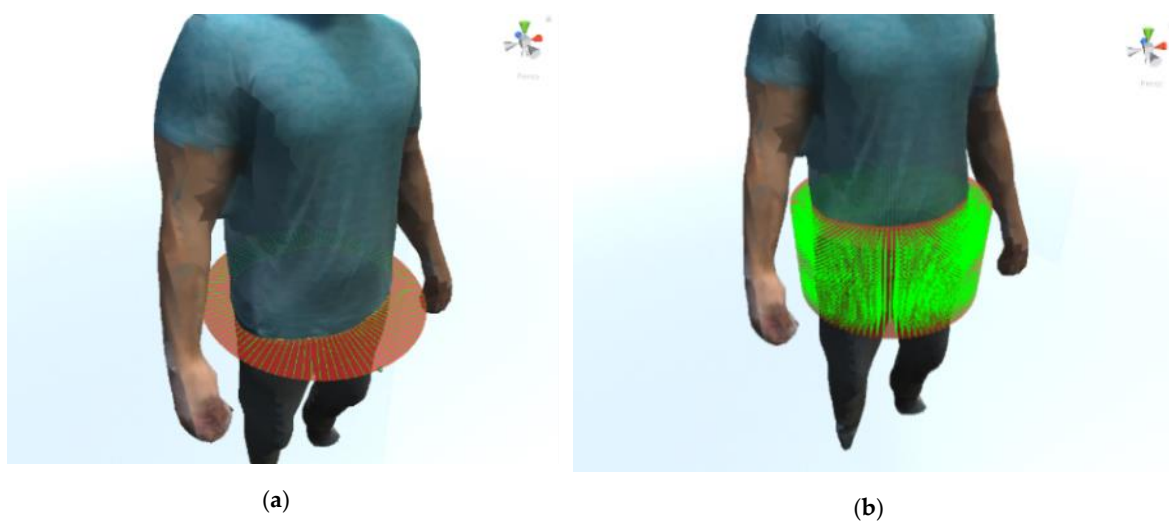

(**a**)          (**b**)

**Figure 9.** Rays perpendicularly launched along the circumferential chord of the circle storing the points of collision with the mesh (**a**). Rays perpendicularly launched along the circumferential chord of the circles inside the cylinder storing the points of collision with the mesh (**b**).

2.3.2. Estimation of Area and Volume

For the estimation of the area enclosed in the perimeter of a section of a 3D model, the set of points obtained in the calculation of the perimeter has been used. From this set of points, a pivot point has been selected for triangulating the area enclosed in the perimeter (see Figure 10). That point will be the middle of the circle. The sum of the areas of the enclosed triangles estimates the sectioned area. The use of a single point as the pivot ensures that these areas do not overlap since the figure is not convex. This method is valid for this application since all body sections used are convex and the intersection lies in a plane. As with the relationship between the number of projected rays and the precision of the perimeter estimation, the number of triangles used is directly related to the precision of the area measurement and its temporal cost, as shown in the experimental section.

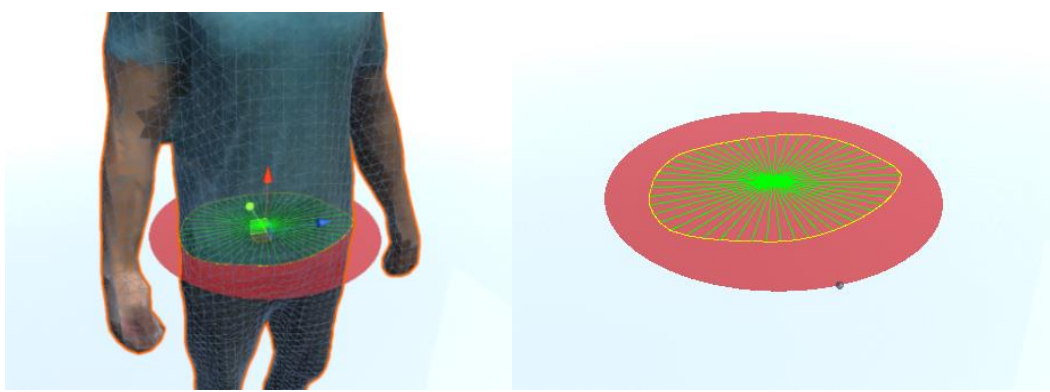

**Figure 10.** Triangulation of the area enclosed in the perimeter.

As mentioned in Section 2.3.1, for the calculation of volumes selected from 3D mesh-based models, a cylinder is used that intersects the volume to be measured, depicted in Figure 9b. The upper and lower circles of the cylinder determine the upper and lower planes of the 3D volume to be measured. The method used to estimate the volume is based on the use of the triangulation method for the area calculation of a section, by iterating from the upper to the lower circle and assuming a pre-set height "h" for each of the sections.

## 3. Results

In this section, the results achieved with the different proposed methods are reported. In the first subsection, we provide quantitative data about the accuracy achieved by the acquisition system in terms of relative and absolute error of the estimated measurements with respect to known measures (ground truth). We provide experimentation on both synthetic and real objects. The experiments on synthetic objects isolate the error corresponding to the measurement methods (presented in Section 2.3) since the objects are not scanned by the system. The experimentation over real objects provides the error corresponding to the acquisition system, i.e., camera noise, calibration, and reconstruction. The second subsection provides evidence of the quality achieved with the realistic visualizations on the body models incorporating texture.

*3.1. Quantifying the Accuracy of the Method for Measuring Scanned 3D Models*

The following section uses a battery of experiments to quantify the accuracy of the measuring methods. As explained previously, the complete process includes both the scanning of objects/bodies to obtain the corresponding mesh-based 3D models and the methods for obtaining 1D (perimeter), 2D (area) and 3D (volume) measurements the of selected sections of these models. Given the complexity of the complete process and the interest in differentiating where the error comes from (scanning or measuring) it is proposed to carry out both experiments with synthetic and real objects. In this way, the experiments with synthetic objects will be used to estimate the error of the methods for obtaining measurements, while in the experimentation with real objects, the error of the complete process is measured, being both the scanning error and the error of the methods for extracting measurements.

3.1.1. Experimentation with Synthetic 3D Models

The objective of the experimentation using synthetic models is to quantify the error of the measuring methods explained in Section 3. These 3D models have been designed with known sizes so that the objective is to retrieve that size as accurately as possible with our methods and quantify the error.

Synthetic experimental setup: The following 3D models have been designed using Blender. Table 1 shows the measurements of the objects in cm:

**Table 1.** Synthetic objects and their 1D, 2D and 3D measurements in cm.

| Object | Perimeter (1D) | Area (2D) | Volume (3D) |
|---|---|---|---|
| Cube 1 | 60.00 | 225.00 | 3,375.00 |
| Cube 2 | 200.00 | 2,500.00 | 125,000.00 |
| Cylinder 1 | 300.00 | 5,000.00 | 196,349.54 |
| Cylinder 2 | 200.00 | 2,500.00 | 98,174.77 |
| Cone | 161.80 | 1,250.00 | 32,724.92 |
| Pyramid | 97.00 | 450.00 | 9,000.00 |

Table 2 shows the perimeter estimates using the method described in Section 3 and the relative error of the different objects. The estimation has been calculated by varying the number of rays projected onto the mesh from $10^2$ to $10^5$. It is observed that the average relative error decreases as the number of projected rays increases. It is observed that from $10^4$ the increase in the number of rays does not improve the estimation.

**Table 2.** Perimeter estimates in cm for different numbers of rays and relative error (Rel. $\varepsilon$).

| Object | $10^2$ rays | Rel. $\varepsilon$ | $10^3$ rays | Rel. $\varepsilon$ | $10^4$ rays | Rel. $\varepsilon$ | $10^5$ rays | Rel. $\varepsilon$ |
|---|---|---|---|---|---|---|---|---|
| Cube 1 | 58.47 | 0.025500 | 59.95 | 0.000833 | 59.99 | 0.000167 | 59.93 | 0.001167 |
| Cube 2 | 194.86 | 0.025700 | 199.84 | 0.000800 | 199.99 | 0.000050 | 199.87 | 0.000650 |
| Cylinder 1 | 293.71 | 0.020967 | 299.11 | 0.002967 | 299.86 | 0.000467 | 299.44 | 0.001867 |
| Cylinder 2 | 194.72 | 0.026400 | 199.57 | 0.002150 | 199.84 | 0.000800 | 199.62 | 0.001900 |
| Cone | 157.59 | 0.026020 | 160.14 | 0.010260 | 160.49 | 0.008096 | 159.35 | 0.015142 |
| Pyramid | 93.71 | 0.033918 | 95.81 | 0.012268 | 96.00 | 0.010309 | 95.88 | 0.011546 |
| **Average $\varepsilon$** | | **0.026417** | | **0.004880** | | **0.003315** | | **0.005379** |

Table 3 shows the area estimates for the different objects and the relative error. We provide the estimation varying the number of rays projected onto the mesh from $10^2$ to $10^5$. It is observed that the average relative error decreases as the number of projected rays increases. As with the perimeter estimations, it is observed that from $10^4$ rays on, the estimation does not improve.

**Table 3.** Area estimates (cm$^2$) for different numbers of rays and relative error (%).

| Object | $10^2$ rays | Rel. $\varepsilon$ | $10^3$ rays | Rel. $\varepsilon$ | $10^4$ rays | Rel. $\varepsilon$ | $10^5$ rays | Rel. $\varepsilon$ |
|---|---|---|---|---|---|---|---|---|
| Cube 1 | 224.58 | 0.00187 | 225.01 | 0.000044 | 225.03 | 0.00013 | 225.13 | 0.000578 |
| Cube 2 | 2,495.36 | 0.00186 | 2,500.04 | 0.000016 | 2,500.07 | 0.00003 | 2,500.61 | 0.000244 |
| Cylinder 1 | 4,980.81 | 0.00384 | 4,992.94 | 0.001412 | 4,993.92 | 0.00122 | 4,990.69 | 0.001862 |
| Cylinder 2 | 2,491.79 | 0.00328 | 2,496.41 | 0.001436 | 2,496.54 | 0.00138 | 2,494.01 | 0.002396 |
| Cone | 1,247.78 | 0.00178 | 1,246.22 | 0.003024 | 1,247.16 | 0.00227 | 1,245.64 | 0.003488 |
| Pyramid | 448.45 | 0.00344 | 449.78 | 0.000489 | 450.15 | 0.00033 | 451.43 | 0.003178 |
| **Average $\varepsilon$** | | **0.00268** | | **0.00107** | | **0.00089** | | **0.00196** |

Table 4 shows the estimated volume for the different objects and the relative error. We provide the estimation varying the number of rays projected onto the mesh from $10^2$ to $10^5$. It is observed that the average relative error decreases as the number of projected rays increases. The value of "h" used for the sections of each circle was 1 cm.

As a conclusion, we can state that the relative error attributable to the measuring methods is very low, on the order of 0.005 cm. Furthermore, we can affirm that the increase in the number of rays projected decreases the error until $10^4$ rays, where the error converges. However, given the increase in the temporary cost of using more rays and the low error, it seems desirable to not use too many rays.

**Table 4.** Volume estimates ($cm^3$) for different numbers of rays and relative error (%).

| Object | $10^2$ rays | Rel. ε | $10^3$ rays | Rel. ε | $10^4$ rays | Rel. ε | $10^5$ rays | Rel. ε |
|---|---|---|---|---|---|---|---|---|
| Cube 1 | 3,368.75 | 0.00185 | 3,375.07 | 0.000021 | 3,375.15 | 0.00004 | 3,376.2 | 0.000356 |
| Cube 2 | 124,768.49 | 0.00185 | 125,002.1 | 0.000017 | 125,006.82 | 0.00005 | 125,033.98 | 0.000272 |
| Cylinder 1 | 194,287.47 | 0.01050 | 194,832.98 | 0.007724 | 194,716.61 | 0.00832 | 195,587.54 | 0.003881 |
| Cylinder 2 | 97,385.65 | 0.00804 | 97,623.71 | 0.005613 | 97,526.31 | 0.00661 | 97,431.41 | 0.007572 |
| Cone | 32,379 | 0.01057 | 32,507.18 | 0.006654 | 32,511.76 | 0.00651 | 32,671.93 | 0.001619 |
| Pyramid | 8,970.9 | 0.00323 | 9,003.78 | 0.000420 | 9,006.89 | 0.00077 | 9,009.91 | 0.001101 |
| **Average ε** | | **0.00601** | | **0.00341** | | **0.00372** | | **0.00247** |

3.1.2. Experimentation with Real 3D Objects

The aim of the experimentation with real objects/bodies is to measure the error for the entire scanning and measurement system. The sizes of the real objects/bodies are estimated by manual procedures and their 3D models have been obtained by the scanning system detailed in Section 2. Since the errors introduced by the measurement methods (Section 3) have been estimated to be very low, the error studied in this section will be mostly due to the scanning system. Since, in the previous section, a suitable number of rays was found to be $10^4$, we use it in this section.

Real experimental setup: For the real experimentation we have used the following 3D models: a cube and different parts of the body. We have measured these models by manual procedures. 2D and 3D real measures are only available for the cube.

Table 5 shows the real measurements of the cube (perimeter, area and volume) in (cm, $cm^2$ and $cm^3$) in the columns R.1D, R.2D and R.3D, and it shows the estimates in the columns E.1D, E.2D and E.3D, obtained with the proposed methods with the 13-camera system. The relative errors for these measurements in relation to the real measurements obtained by manual procedures is also observed (Rel. ε1, Rel. ε2 and Rel. ε3), whose average relative error is 0.873%.

Table 6 shows the real mono-dimensional measurements using conventional meters for different parts of the body (R.1D) and the estimated measurements extracted using the proposed methods (E.1D), with both 13- and 8-camera systems. The absolute and relative errors for these measurements in relation to the real ones are also provided (Abs. ε, Rel. ε) (13 Cam, 8 Cam). It is observed that the average absolute error for perimetral measurements is 0.984 cm and the average relative error is 2.78% for the 13-camera system. In order to compare the accuracy, the 8-camera system is positioned at the same position as the 13-camera-system. For this reason, the 8-camera system does not provide enough information to extract measurements on certain parts of the body (calf, elbow, wrist, forearm, hand/knuckles), since they are out of field of view. Hence, the measurements of these parts are not shown in the table. The rest of the body parts that are visible to the 8-camera system show an average absolute error of 2.342 cm and an average relative error of 4.80%, higher than the 13-camera system. A simple series of five manual measurements of the calf with conventional meters has also been made. The five measurements were taken by different people with the same instructions. We observed an average absolute error of 0.64 cm and a relative error of 1.905%, which is a similar order of error to that provided by the 13-camera acquisition system.

**Table 5.** Cube real measurements (R.1D, R.2D, R.3D), their estimates in cm (E.1D, E.2D, E.3D), and their relative errors.

| Object | R.1D | E.1D | Rel. ε1 | R.2D | E. 2D | Rel. ε2 | R.3D | E.3D | Rel. ε3 |
|---|---|---|---|---|---|---|---|---|---|
| Cube 1 | 100 | 99.68 | 0.0032 | 625.00 | 632.09 | 0.011 | 15,625 | 15,805.6 | 0.012 |

**Table 6.** Real body part perimeter measurements (R.1D) with both 13- and 8-camera systems, and their estimates (E.1D) in cm. Their relative and absolute errors.

| Body | Real. 1D | E.1D 13 Cam | E.1D 8 Cam | Abs. ε 13 Cam | Rel. ε 13 Cam | Abs. ε 8 Cam | Rel. ε 8 Cam |
|---|---|---|---|---|---|---|---|
| Calf | 32 | 32.52 | - | 0.52 | 0.016250 | - | - |
| Quadriceps | 44.50 | 44.77 | 41.71 | 0.27 | 0.006067 | 2.79 | 0.062697 |
| Waist | 74 | 75.15 | 72.88 | 1.15 | 0.015541 | 1.12 | 0.015135 |
| Hip | 84.5 | 85.20 | 84.68 | 0.70 | 0.008284 | 0.18 | 0.002130 |
| Elbow | 24.80 | 24.97 | - | 0.17 | 0.006855 | - | - |
| Wrist | 16 | 16 | - | 0.00 | 0.000000 | - | - |
| Biceps | 25.80 | 26.49 | 26.32 | 0.69 | 0.026744 | 0.52 | 0.020155 |
| Forearm | 22 | 21.58 | - | 0.42 | 0.019091 | - | - |
| Forehead | 56.70 | 56.48 | 63.21 | 0.22 | 0.003880 | 6.51 | 0.114815 |
| Neck | 34.10 | 36.48 | 35.29 | 2.38 | 0.069795 | 1.19 | 0.034897 |
| Ankle | 20.80 | 20.96 | 21.33 | 0.16 | 0.007692 | 0.53 | 0.025481 |
| Chest (low rib) | 78.50 | 78.20 | 79.01 | 0.30 | 0.003822 | 0.51 | 0.006497 |
| Shoulders | 107 | 106.59 | 110.9 | 0.41 | 0.003832 | 3.90 | 0.036449 |
| Knee | 36.40 | 31.33 | 42.07 | 5.07 | 0.139286 | 5.67 | 0.155769 |
| Hand (knuckles) | 25.40 | 27.70 | - | 2.30 | 0.090551 | - | - |
| **Average ε** | | | | **0.984** | **0.027846** | **4.66** | **0.176652** |

Finally, our proposal has been quantitively compared with state-of-the-art methods. Tables 7 and 8 show the absolute (millimeters) and relative error (percentage), respectively, for the most-used anthropometric measurements that provide detailed information about the body shape: specifically, the chest, waist and hip girths. According to the absolute error (Table 7), our method achieves a very high accuracy, obtaining only an absolute error of 3 mm for the chest, 11.5 mm for the waist, and 7 mm for the hip girth. Compared to the state of the art, our method achieves the best results for the chest and waist, providing 5.95 mm and 4.3 mm, respectively: less absolute error than the state of the art ([34,35]). Regarding the hip girth, the method of Human Mesh Recovery HMR [34] is 0.85 mm better than ours. In any case, for these measurements, on average our method has an error below 1 cm (7.16 mm) whereas the method in [35] provides 1.25 cm of absolute error.

**Table 7.** Comparative results of body part perimeters according to the absolute error (expressed in millimeters). Lower values are better. The best results are in bold.

| Estimation method | Chest | Waist | Hip | Average |
|---|---|---|---|---|
| KPhub-I [34] | 12.92 | 23.72 | 8.43 | 15.02 |
| SMPLify [34] | 8.95 | 24.97 | 12.10 | 15.34 |
| HMR [34] | 43.39 | 16.01 | **6.15** | 21.85 |
| [35] | 12.5 | 15.8 | 9.3 | 12.53 |
| [36] | 22.8 | 24 | 20 | 22.27 |
| [37] | 92.8 | 118.3 | 68.7 | 93.27 |
| Ours | **3.0** | **11.5** | 7.0 | **7.16** |

The comparison of absolute error provides interesting results as a measurement system, but it does not take into account the size of the bodies (or the parts) to be compared. Thus, state-of-the-art

415 of 20works often use relative error measurement to allow comparison with other methods. Table 8 shows the relative error for state-of-the-art methods compared to our proposal. Again, our method achieves very high accuracy, providing the best results for chest and hip in this comparison (0.38 % and 0.82%, respectively). In this case, the best results for the waist girth are achieved in the work of Uhm et al. [38], being the best for the state-of-the-art method (not considering our method). However, this method uses a color-IR marker-based vest with markers to measure these parameters, whereas the others use single view or a set of views to measure the body. In any case, for these measurements, on average our method has an error below 1% (0.92 %), whereas the method in [38] provides 1.49% of the relative error. For this comparison, we can conclude that our method outperforms the state of the art.

**Table 8.** Comparative results of body part perimeters according to the relative error (expressed in percentage). Lower values are better. The best results are in bold.

| Estimation method | Chest | Waist | Hip | Average |
|---|---|---|---|---|
| [38] | 1,676 % | **1.52 %** | 1.29 % | 1.49 % |
| PRA [39] | 4.87 % | 4.30 % | 5.63 % | 4.93 % |
| EIR [39] | 5.02 % | 4.74 % | 4.57 % | 4.78 % |
| P2P [39] | 5.06 % | 4.30 % | 4.16 % | 4.51 % |
| [40] | 11.60 % | 10.97 % | N/A | 11.28 % |
| [41] | 4.76 % | 4.22 % | 6.46 % | 5.15 % |
| [42] | 1.49 % | 2.78 % | 1.81 % | 2.03 % |
| Basic [43] | 4.70 % | 9.01 % | 1.10 % | 4.93 % |
| Camera [43] | 4.85 % | 9.10 % | 1.20 % | 5.05 % |
| Styku [43] | 2.50 % | 6.25 % | 2.40 % | 3.72 % |
| [44] | 2.11 % | 4.66 % | 4.31 % | 3.69% |
| **Ours** | **0.38 %** | 1.55 % | **0.82 %** | **0.92 %** |

*3.2. Body Model Visualization*

In relation to the visualization results, below we show different models obtained by the system incorporating the projected texture. As can be seen in Figures 11 and 12, the quality of the images provided by the system is adequate for the application level, although we are studying the introduction of techniques to improve their realism. We show demo videos for different models at [32].

The following figures shows different views of the result of the texture projection on the mesh-based model.

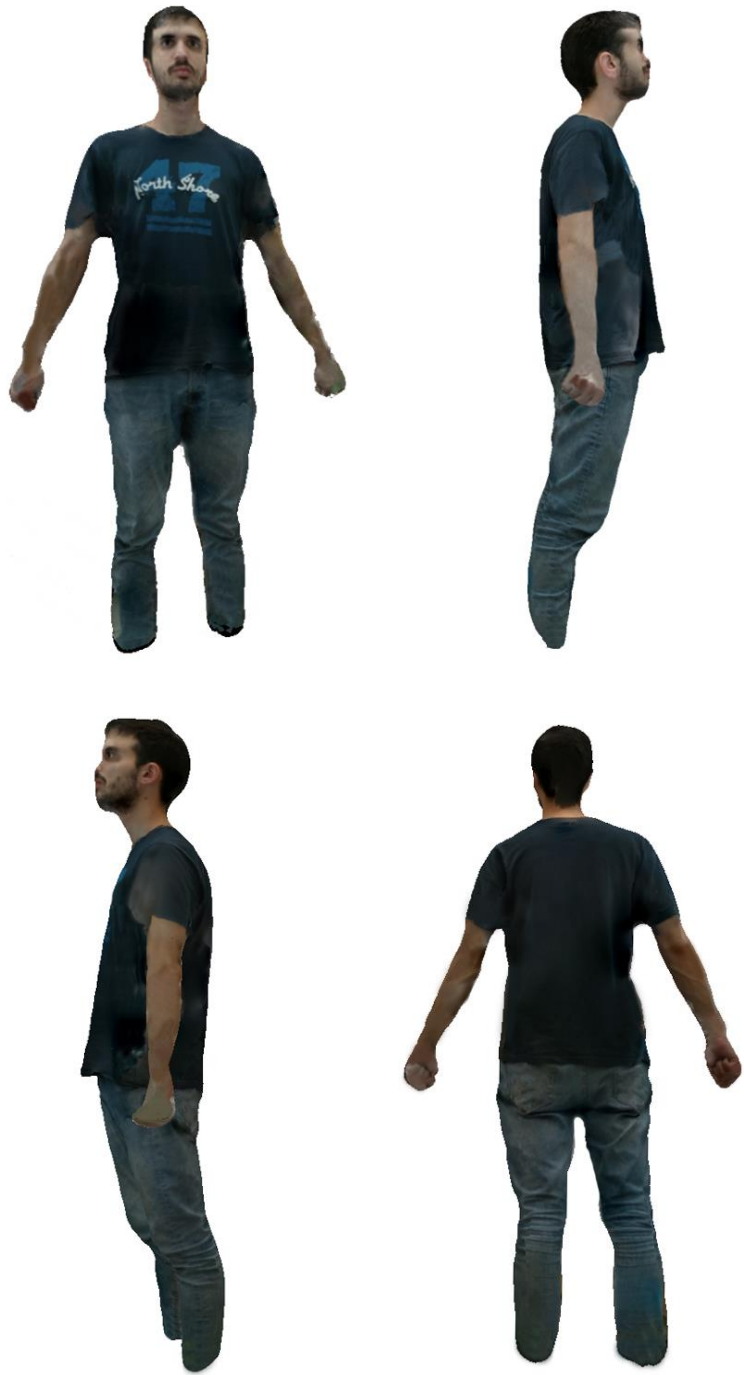

**Figure 11.** Different views of the same textured body model.

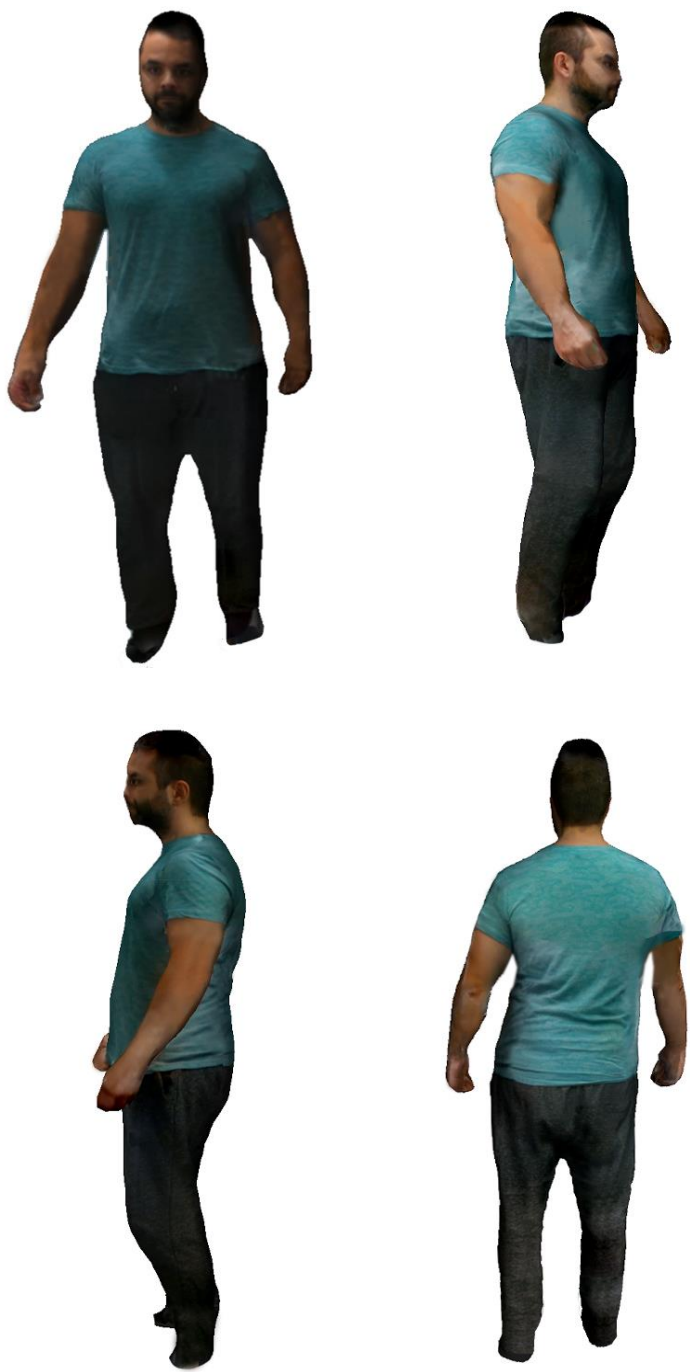

**Figure 12.** Different views of the same textured body model.

## 4. Conclusions

This work focuses its contribution on developing technologies for health improvement. Specifically, 3D cameras (RGB-D) and virtual reality technologies are intended to improve obesity treatment. The developed framework provides the opportunity to investigate the positive effect of realistic 3D representations on the body awareness and psychological well-being of the individual and to study problems related to adherence to dietetic-nutritional treatment. Moreover, the proposed methodology allows a 3D visual model of the human body to be obtained over time to analyze the morphological progress due to changes resulting from obesity treatments, with the possibility of obtaining precise body measures and analyzing the evolution of these measures (1D, 2D and 3D) over time (4D). The 3D models were obtained from the acquisition of multiple views through a network of RGB-D cameras. These views were filtered and aligned to obtain a mesh model on which the texture was projected generating realistic 3D models. Sequences of realistic 3D models allow the generation of the 4D models used in the visualizations, which we hope will provide a powerful method to improve adherence to obesity treatments. Different scientific challenges in the area of computer vision were addressed using well-known state of the art methods: improving the quality of point clouds, obtaining a precise 3D mesh-based body model and projecting textures over the mesh generating realistic 3D representation. In addition, new methods have been proposed to address the problem of calibrating RGB-D camera networks using 3D markers, as well as obtaining 1D, 2D and 3D measures from the mesh and its evolution over time (4D). The quantitative comparison of our method with respect to the state of the art shows that we achieve a very high accuracy, outperforming the methods in the literature. Furthermore, from a technological point of view, a framework based on low-cost, adaptable and portable technologies and intuitive environments based on realistic 3D representations has been designed. The potential of these technologies, together with the purpose of improving the treatment of obesity, allows high social and economic impact to be possible. The system is planned to be installed in three primary health care centers in order to obtain and study the psychological results. Subsequently, it is planned to transfer the operation to health institutions in the area of Alicante (Spain).

**Author Contributions:** conceptualization and methodology, A.F.G., J.A.L.; software, J.M.C.Z., N.G.D.; validation, M.S.C.; writing—review, A.F.G., J.A.L.; supervision, R.B.F. All authors have read and agreed to the published version of the manuscript.

**Funding:** This work has been partially funded by the Spanish Government TIN2017-89069-R grant supported with Feder funds.

**Conflicts of Interest:** The authors declare no conflict of interest.

## References

1. Boraxbekk, C.J; Stomby, A., Ryberg, M., Lindahl, B., Larsson, C., Nyberg, L.; Olsson, T. Diet-induced weight loss alters functional brain responses during an episodic memory task. *Obes. Facts* **2015**, *8*, 261–272, doi:10.1159/000437157.
2. Drigny, J.; Gremeaux, V.; Dupuy, O.; Gayda, M.; Bherer, L.; Juneau, M.; Nigam, A. Effect of interval training on cognitive functioning and cerebral oxygenation in obese patients: A pilot study. *J. Rehabil. Med.* **2014**, *46* 1050–1054, doi:10.2340/16501977-1905.
3. Lehnert, T.; Sonntag, D.; Konnopka, A.; Riedel-Heller S.; König, H.-H. Economic costs of overweight and obesity. *Best Pract. Res. Clin. Endocrinol. Metab.* **2013**, 27, 105–115, doi:10.1016/j.beem.2013.01.002.
4. Withrow, D.; Alter, D.A. The economic burden of obesity worldwide: A systematic review of the direct costs of obesity. *Obes. Rev.* **2011**, doi:10.1111/j.1467-789X.2009.00712.x.
5. Sicras-Mainar, A.; Gil, J.; Mora, T.; Ayma, J. Prevalencia e impacto económico de la obesidad en adultos durante el periodo 2003–2010. *Med. Clin*. **2012**, doi:10.1016/j.medcli.2012.02.006.
6. De Geest, S.; Sabaté, E. Adherence to long-term therapies: Evidence for action. *Eur. J. Cardiovasc. Nurs*. **2003**, doi:10.1016/S1474-5151(03)00091-4.
7. Boeka, A.; Lokken, K. Neuropsychological performance of a clinical sample of extremely obese individuals. *Arch. Clin. Neuropsychol.* **2008**, doi:10.1016/j.acn.2008.03.003.
8. Roseman, M.G.; Riddell, M.C.; Haynes, J.N. A content analysis of kindergarten-12th grade school-based nutrition interventions: Taking advantage of past learning. *J. Nutr. Educ. Behav.* **2011**, doi:10.1016/j.jneb.2010.07.009.
9. Ajie, W.N.; Chapman-Novakofski, K.M. Impact of computer-mediated, obesity-related nutrition education interventions for adolescents: A systematic review. *J. Adolesc. Heal.* **2014**, doi:10.1016/j.jadohealth.2013.12.019.
10. Ferrer-Garcia, M.; Gutiérrez-Maldonado, J.; Riva, G. Virtual reality based treatments in eating disorders and obesity: A review. *J. Contemp. Psychother.***2013**, *43*, 207–221, doi:10.1007/s10879-013-9240-1.
11. Kuzmar, I.; Rizo, M.; Cortés-Castell, E. Adherence to an overweight and obesity treatment: How to motivate a patient. *PeerJ* **2014**, doi:10.7717/peerj.495.

12. Fuster-Guilló; Azorín-López; Zaragoza; Pérez; Saval-Calvo; and Fisher. 3D technologies to acquire and visualize the human body for improving dietetic treatment. *Proceedings* **2019**, doi:10.3390/proceedings2019031053.
13. He, Q.; Ji, Y.; Zeng, D.; Zhang, Z. Volumeter: 3D human body parameters measurement with a single kinect. *IET Comput. Vis.* **2018**, *12*, 553–561.doi:10.1049/iet-cvi.2017.0403.
14. Treleaven, P.; Wells, J. 3D body scanning and healthcare applications. *Computer* **2007**, doi:10.1109/MC.2007.225.
15. Lin, J.-D.; Chiou, W.-K.; Weng, H.-F.; Tsai, Y.-H.; Liu, T.-H. Comparison of three-dimensional anthropometric body surface scanning to waist-hip ratio and body mass index in correlation with metabolic risk factors. *J. Clin. Epidemiol.* **2002**, *55*, 757–766, doi:10.1016/S0895-4356(02)00433-X.
16. Alldieck, T.; Magnor, M.A.; Xu, W.; Theobalt, C.; Pons-Moll, G. Detailed Human Avatars from Monocular Video. 2018. Available Online: https://www.semanticscholar.org/paper/Detailed-Human-Avatars-from-Monocular-Video-Alldieck-Magnor/07377c375ac76a34331c660fe87ebd7f9b3d74c4. (accessed on 10 October 2018).
17. Yu, T.; Zheng, Z., Guo, K., Zhao, J., Dai, Q., Li, H.; Liu, Y. Doublefusion: Real-time capture of human performances with inner body shapes from a single depth sensor. In Proceedings of the 2018 Conference on Computer Vision and Pattern Recognition, Salt Lake City, UT, USA. 18–23 June 2018, doi:10.1109/CVPR.2018.00761.
18. Fit 3D Body Scanners. Available Online: https://fit3d.com/ (accessed on 14 July 2019).
19. Naked—The World's First Home Body Scanner. Available Online: https://nakedlabs.com/ (accessed on 14 July 2019).
20. Villena-Martínez, V.; Fuster-Guilló, A.; Azorín-López, J.; Saval-Calvo, M.; Mora-Pascual, J.; Garcia-Rodriguez, J.; Garcia-Garcia, A. A quantitative comparison of calibration methods for RGB-D sensors using different technologies. *Sensors* **2017**, *17*, 243,.doi:10.3390/s17020243.
21. Hussein, M.; Nätterdal, C. The benefits of virtual reality in education—A comparision study. Bachelor Thesis, University of Gothenburg, Gothenburg, Sweden, **2015**.
22. Saval-Calvo, M.; Azorin-Lopez, J.; Fuster-Guillo, A.; Garcia-Rodriguez, J. Three-dimensional planar model estimation using multi-constraint knowledge based on k-means and RANSAC. *Appl. Soft Comput. J.* **2015**, doi:10.1016/j.asoc.2015.05.007.
23. Saval-Calvo, M.; Azorin-Lopez, J.; Fuster-Guillo, A.; Mora-Mora, H. μ-MAR: multiplane 3D marker based registration for depth-sensing cameras. *Expert Syst.* **2015**, *42*, 9353–9365.
24. Fischler M.A.; Bolles, R.C. Random sample consensus: A paradigm for model fitting with applications to image analysis and automated cartography. *Commun. ACM* **1981**, *24*, 381–395, doi:10.1145/358669.358692.
25. PCL Team. "Point Cloud Library (PCL): pcl::MedianFilter< PointT > Class Template Reference," 2013. http://docs.pointclouds.org/1.7.1/classpcl_1_1_median_filter.html (accessed on 27 May 2019).
26. PCL Team. "Point Cloud Library (PCL): pcl::BilateralFilter< PointT > Class Template Reference," 2019. http://docs.pointclouds.org/trunk/classpcl_1_1_bilateral_filter.html (accessed on 27 May 2019).
27. PCL Team. "Point Cloud Library (PCL): pcl::StatisticalOutlierRemoval< PointT > Class Template Reference," 2013. http://docs.pointclouds.org/1.7.1/classpcl_1_1_statistical_outlier_removal.html (accessed on 27 May 2019).
28. Rusu, R.B. Documentation—Point Cloud Library (PCL). Available Online: http://pointclouds.org/documentation/tutorials/normal_estimation.php (accessed on 19 May 2019).
29. Saval-Calvo, M.; Azorín-López, J.; Fuster-Guilló, A. Model-based multi-view registration for RGB-D sensors. *Lect. Notes Comput. Sci.* **2013**, doi:10.1007/978-3-642-38682-4_53.
30. Kazhdan, M.; Bolitho, M.; Hoppe, H. Poisson Surface Reconstruction. 2006. Available online: http://hhoppe.com/poissonrecon.pdf. (accessed on 28 May 2019.)
31. Callieri, M.; Cignoni, P.; Corsini, M.; Scopigno, R. Masked photo blending: Mapping dense photographic data set on high-resolution sampled 3D models. *Comput. Graph.* **2008**, *32*, 464–473, doi:10.1016/j.cag.2008.05.004.
32. Media—Tech4Diet: Project TIN2017-89069-R Spanish State Research Agency (AEI). 4D Modelling and Visualization of the Human Body to Improve Adherence to Dietetic-Nutritional Intervention of Obesity, 2019. Available online: http://tech4d.dtic.ua.es/media/ (accessed on 06 November 2019).
33. Unity. *Unity User Manual (2018.3) - Unity Manual*; 2018. Available online: https://docs.unity3d.com/2018.3/Documentation/Manual/ (accessed on 27 May 2019)

34. Xu, Z.; Chang, W.; Zhu, Y.; Le, D.; Zhou, H.; Zhang, Q. Building high-fidelity human body models from user-generated data. *IEEE Trans. Multimed*. **2020**, 1, doi:10.1109/tmm.2020.3001540.
35. Smith, B.M.; Chari, V.; Agrawal, A.; Rehg, J.M.; Sever, R. Towards accurate 3D human body reconstruction from silhouettes. In Proceedings of the 2019 International Conference on 3D Vision (3DV). 2019, 279–288, doi:10.1109/3DV.2019.00039.
36. Dibra, E.; Jain, H.; Oztireli, C.; Ziegler, R.; Gross, M. HS-Nets: Estimating human body shape from silhouettes with convolutional neural networks. In Proceedings of the 2016 fourth international conference on 3D vision (3DV).2016, 108–117, doi:10.1109/3DV.2016.19.
37. Kanazawa, A.; Black, M.J.; Jacobs, D.W.; Malik, J. End-to-End recovery of human shape and pose. In Proceedings of the IEEE Conference on Computer Vision and Pattern Recognition. 2018, 7122–7131.doi:10.1109/CVPR.2018.00744.
38. Uhm, T.; Park, H.; Park, J. Il. Fully vision-based automatic human body measurement system for apparel application. Meas. J. Int. Meas. Confed. 2015, 61, 169–179, doi:10.1016/j.measurement.2014.10.044.
39. Albances, X.; Binungcal, D.; Nikko Cabula, J.; Cajayon, C.; Cabatuan, M. RGB-D camera based anthropometric measuring system for barong tagalog tailoring. In Proceedings of the 2019 IEEE 11th International Conference on Humanoid, Nanotechnology, Information Technology, Communication and Control, Environment, and Management (HNICEM), Laoag, Philippines, 29 November–1 December 2019, doi:10.1109/HNICEM48295.2019.9072869.
40. Non-Contact Human Body Parameter Measurement Based on Kinect Sensor human body measurement camera rgbd hips chest relative error. *IOSR J. Comput. Eng.* **2017**, *19*, 80–85.
41. Adikari, S.B.; Ganegoda, N.C.; Meegama, R.G.N.; Wanniarachchi, I.L. Applicability of a Single Depth Sensor in Real-Time 3D Clothes Simulation: Augmented Reality Virtual Dressing Room Using Kinect Sensor. *Adv. Hum.Comput. Interact.* **2020**, *2020*, 1314598, doi:10.1155/2020/1314598.
42. Xu, H.; Li, J.; Li, J.; Lu, G. Prediction of anthropometric data based on ladder network. In Proceedings of the 2019 Chinese Automation. Congress. (CAC) , Hangzhou, China, 22–24, November, 2019; pp. 512–517, doi:10.1109/CAC48633.2019.8997368.
43. Peeters T.; et al. A comparative study between three measurement methods to predict 3D body dimensions using shape modelling. *Adv. Intell. Syst. Comput.* **2020**, *975*, 464–470, doi:10.1007/978-3-030-20216-3_43.
44. Gan, X.Y., Ibrahim, H. and Ramli, D.A. A simple vision based anthropometric estimation system using webcam. *J. Phys. Conf. Ser.* **2020**, *1529*, 022067, doi:10.1088/1742-6596/1529/2/022067.